\documentclass[sigconf,natbib=true,anonymous=false]{acmart}
\AtBeginDocument{%
  \providecommand\BibTeX{{%
    \normalfont B\kern-0.5em{\scshape i\kern-0.25em b}\kern-0.8em\TeX}}}

\usepackage{booktabs} 

\usepackage[english]{babel}
\usepackage{moresize}
\usepackage{amsmath}
\usepackage{algorithmic}
\usepackage{balance}
\usepackage{comment}
\usepackage{paralist}
\usepackage{bm}
\usepackage{pgfplots}
\usetikzlibrary{pgfplots.dateplot}

\usepackage{flushend}
\usepackage[english]{babel}
\usepackage[latin1]{inputenc}
\usepackage{mathrsfs}
\usepackage{graphicx}

\usepackage{amssymb}
\usepackage{amsfonts}
\usepackage{url}
\usepackage{longtable}
\usepackage{rotating}
\usepackage{multirow}
\usepackage{mathrsfs}
\usepackage{subfigure}
\usepackage{enumitem}
\usepackage[linesnumbered,algoruled,boxed,lined]{algorithm2e}
\usepackage{adjustbox}
\usepackage{hyperref}
\usepackage{pgfplots}
\usetikzlibrary{pgfplots.dateplot}
\usepackage{filecontents}
\usepackage{threeparttable}
\definecolor{tblue}{RGB}{31,119,180}
\definecolor{torange}{RGB}{255,127,14}
\definecolor{tgreen}{RGB}{44,160,44}
\definecolor{tred}{RGB}{214,39,40}
\definecolor{tpurple}{RGB}{148,103,189}

\newcommand{\hide}[1]{} 

\newcommand{\ie}{\textit{i}.\textit{e}.}
 
\newcommand{\wrt}{\textit{w}.\textit{r}.\textit{t}}

\def\model{STExplainer}
\newcommand{\post}{\emph{post-hoc}} 
\newcommand{\intr}{\emph{intrinsic}} 

\copyrightyear{2023}
\acmYear{2023}
\setcopyright{acmlicensed}\acmConference[CIKM '23]{Proceedings of the 32nd ACM International Conference on Information and Knowledge Management}{October 21--25, 2023}{Birmingham, United Kingdom}
\acmBooktitle{Proceedings of the 32nd ACM International Conference on Information and Knowledge Management (CIKM '23), October 21--25, 2023, Birmingham, United Kingdom}
\acmPrice{15.00}
\acmDOI{10.1145/3583780.3614871}
\acmISBN{979-8-4007-0124-5/23/10}

\ccsdesc[500]{Information systems~Spatial-temporal systems}
\ccsdesc[500]{Information systems~Data mining}
\ccsdesc[500]{Computing methodologies}
\ccsdesc[500]{Computing methodologies~Neural networks}

\keywords{Spatio-Temporal Data Mining; Graph Neural Networks; Urban Computing; Explainable AI}

\begin{document}

\title{Explainable Spatio-Temporal Graph Neural Networks}

\author{Jiabin Tang}
\affiliation{
  \institution{University of Hong Kong \\ \country{Hong Kong, China}}
}
\email{jiabintang77@gmail.com}

\author{Lianghao Xia}
\affiliation{
  \institution{University of Hong Kong \\ \country{Hong Kong, China}}
}
\email{aka_xia@foxmail.com}

\author{Chao Huang}
\authornote{Chao Huang is the corresponding author.}
\affiliation{
  \institution{University of Hong Kong \\ \country{Hong Kong, China}}
}
\email{chaohuang75@gmail.com}

\renewcommand{\shortauthors}{Jiabin Tang, Lianghao Xia, \& Chao Huang}

\begin{abstract}
Spatio-temporal graph neural networks (STGNNs) have gained popularity as a powerful tool for effectively modeling spatio-temporal dependencies in diverse real-world urban applications, including intelligent transportation and public safety.
However, the black-box nature of STGNNs limits their interpretability, hindering their application in scenarios related to urban resource allocation and policy formulation. To bridge this gap, we propose an Explainable Spatio-Temporal Graph Neural Networks (\model) framework that enhances STGNNs with inherent explainability, enabling them to provide accurate predictions and faithful explanations simultaneously. Our framework integrates a unified spatio-temporal graph attention network with a positional information fusion layer as the STG encoder and decoder, respectively. Furthermore, we propose a structure distillation approach based on the Graph Information Bottleneck (GIB) principle with an explainable objective, which is instantiated by the STG encoder and decoder. Through extensive experiments, we demonstrate that our \model\ outperforms state-of-the-art baselines in terms of predictive accuracy and explainability metrics (\ie, sparsity and fidelity) on traffic and crime prediction tasks. Furthermore, our model exhibits superior representation ability in alleviating data missing and sparsity issues. The implementation code is available at: \url{https://github.com/HKUDS/\model}.

\end{abstract}



\maketitle
\vspace{-0.05in}
\section{Introduction}
\label{sec:intro}

The accurate spatio-temporal prediction holds great significance in addressing challenges related to transportation management and public safety risk assessment across a wide range of real-world applications. These applications encompass traffic prediction~\cite{DCRNN,DeepST} and crime forecasting~\cite{DeepCrime, ST-SHN}. The primary objective of spatio-temporal prediction is to capture and comprehend the intricate spatial and temporal dynamics present in historical observations, ultimately enabling informed decision-making processes~\cite{STGCN,zhang2023spatial}. By effectively modeling these dynamics, we can facilitate efficient resource allocation, policy formulation, and risk mitigation.

There has been significant research on modeling spatio-temporal signals, resulting in various approaches. Early works often employed Convolutional Neural Networks (CNNs)~\cite{DeepST, ST-ResNet, DMVST-Net, ConvLSTM} for spatial relation mining, while Recurrent Neural Networks (RNNs)~\cite{DCRNN, AGCRN, STDN} and Temporal Convolutional Networks (TCNs)~\cite{GraphWaveNet, DMSTGCN, MTGNN} were utilized for temporal pattern extraction. In recent years, Graph Neural Networks (GNNs)~\cite{STGCN, STMGCN, STGODE} have gained popularity and have been incorporated into state-of-the-art models. Regardless of whether the approach is graph-based or grid-based (the two general categories of spatio-temporal prediction~\cite{DL-Traff}), Spatio-Temporal Graph Neural Networks (STGNNs) demonstrate their strong capability in modeling complex spatio-temporal dependencies~\cite{DSTAGNN,zhang2023automated}. For example, STGNNs leverage the graph structure of the traffic network to capture spatial dependencies among different locations and temporal dependencies across different time intervals. This approach has proven highly effective in various traffic prediction tasks, including traffic flow prediction and traffic speed prediction.

Explainable Artificial Intelligence (XAI)~\cite{LIME} has emerged as a prominent research area, garnering increasing attention. In the context of GNNs, XAI aims to provide transparent and interpretable explanations, improving the trustworthiness of black-box models, and facilitating effective human utilization~\cite{XAI4GNN_survey}. Generally speaking, XAI approaches for GNNs can be broadly categorized into two groups: post-hoc and intrinsic methods. Specifically, post-hoc models offer explanations without relying on GNN inference and can be further classified into instance-level methods~\cite{GNNExplainer, PGExplainer, SubgraphX, GraphMask}, which provide explanations at the individual instance level, and model-level methods~\cite{XGNN}, which provide explanations at the overall model level. In contrast, intrinsic methods leverage the concept of Information Bottleneck (IB)\cite{IB_BB, VIB} to probe the inherent interpretability and generalization of GNNs\cite{GIB, ib-subgraph, GSAT}. Additionally, recent advancements in intrinsic methods~\cite{DIR, CAttn} have addressed the challenges of explainability and handling graph-out-of-distribution scenarios with invariant learning and causal inference.

Despite significant advances in spatio-temporal models and explainability methods for graphs, the field of explainability for spatio-temporal prediction, particularly for spatio-temporal graph neural networks, remains largely unexplored. This creates a critical need for improved deployment of spatio-temporal models in real-world scenarios. Human-interpretable explanations can assist decision-makers in effectively utilizing spatio-temporal models for various downstream tasks, including urban planning, intelligent transportation systems, and emergency resource scheduling. To address this crucial gap, we propose the development of explainability models specifically tailored for STGNNs. However, existing graph explainability models primarily focus on classification tasks, such as \textsc{BA-Shapes}, \textsc{BA-Community}, and \textsc{BA-Cycles}~\cite{GNNExplainer}, and there is currently a lack of ground-truth datasets available for spatio-temporal explainability. Therefore, we need to tackle the following key questions to advance the field of spatio-temporal explainability: \\\vspace{-0.1in}


\noindent \textbf{Q1:} How can the explainability of STGNN be defined?\\\vspace{-0.12in}

\noindent \textbf{Q2:} How to endow STGNN with spatial and temporal explainability to provide insights underlying cross-region and time dependencies? \\\vspace{-0.12in}

\noindent \textbf{Q3:} How to evaluate the performance of STGNN in terms of explainability in the absence of ground-truth labels? \\\vspace{-0.1in}        



\noindent \textbf{Contribution.} 
In this study, we address the aforementioned challenges by presenting Explainable Spatio-Temporal Graph Neural Networks (\model). Our framework offers scalability, interpretability, and generalization capabilities. We achieve this by breaking down the STG into separate spatial and temporal graphs and employing a unified spatio-temporal graph attention network to encode the spatial and temporal dynamics. Furthermore, we integrate spatio-temporal positional information into the STG decoder layer. In our approach, we define explainability as the ability to identify influential spatial and temporal subgraphs that have a significant impact on predictive results. To accomplish this, we propose utilizing the spatio-temporal Graph Information Bottleneck (GIB) with a structure-distilled explainable objective. We employ variational approximation to make the objective tractable and instantiate the variational bounds with our proposed STG encoder and decoder. To evaluate the performance of STGNN in terms of explainability without the availability of ground-truth, we adapt two metrics, \emph{Sparsity} and \emph{Fidelity}, to suit the explainable evaluation of STGNN.


In summary, our work makes the following contributions: 
\begin{itemize}[leftmargin=*]

    \item To the best of our knowledge, we present the first systematic investigation into the explainability of STGNN, specifically focusing on identifying the most influential spatial and temporal subgraphs in relation to the prediction results. \\\vspace{-0.12in}
    

    \item We propose a novel explainable framework \model, which integrates the structure-distilled graph information bottleneck principle with a unified spatio-temporal attentive encoder and decoder to enhance the explainability and generalization of STGNN.\\\vspace{-0.1in}
    

    \item In our proposed \model\ framework, we employ the spatio-temporal graph information bottleneck principle with a structure-distilled explainable objective to control the information flow and characterize it with a unified STGNN. We utilize graph attention networks and a position-aware information fusion layer to encode both interpretable and generalizable STG representations.\\\vspace{-0.1in}
    


    \item We conduct extensive experiments across various settings to evaluate the performance of \model\ in terms of predictive accuracy and explainability. Comparisons over various datasets demonstrate that our model outperforms state-of-the-art baselines.
    
\end{itemize}
            


\section{Preliminaries}
\label{sec:model}

\textbf{Spatio-Temporal Graph Forecasting.}
In Spatio-Temporal Graph (STG) forecasting, we analyze a scenario with $N$ nodes representing regions and $T$ time steps. The spatio-temporal graph $\mathcal{G}=(\mathcal{V}, \mathcal{E}, \mathbf{A}, \mathbf{X})$ is formed, where $\mathcal{V}$ denotes the set of $N$ nodes representing regions, $\mathcal{E}$ represents the edges recorded by the adjacency matrix $\mathbf{A}\in\mathbb{R}^{N\times N}$, and $\mathbf{X}\in\mathbb{R}^{T\times N\times F}$ is the feature matrix associated with attributes like traffic volumes or crime occurrences. Here, $F$ represents the feature dimensions, and $T$ represents the time steps. With these definitions in place, we can formally define the task of spatial-temporal graph forecasting as follows: \\\vspace{-0.12in}


\noindent\textbf{Problem Statement}. 
In STG forecasting, the goal is to learn a predictive function denoted as $f$. This function aims to predict specific attributes of the spatio-temporal graph in the next $L'$ time steps, given the previous $L$ historical observations.
\begin{align}
    \mathbf{Y}_{t:t+L^{\prime}-1} = f(\mathcal{G}(\mathcal{V}, \mathcal{E}, \mathbf{A}, \mathbf{X}_{t-L:t-1}))
\end{align}
\noindent $\mathbf{X}\in \mathbb{R}^{T\times N \times F}$ is the historical observations with $F$ feature dimensions from time step $t-L$ to $t-1$. $\mathbf{Y}\in \mathbb{R}^{L^{\prime}\times N \times F^{\prime}}$ represents the predictions with $F^{\prime}$ feature dimensions for the next $L^{\prime}$ time steps. \\\vspace{-0.12in}

\noindent\textbf{Explainable Graph Neural Networks.}
The research community has recently been captivated by the field of eXplainable Artificial Intelligence (XAI) for Graphs, which focuses on providing reliable and interpretable explanations to enhance the trustworthiness of black-box Graph Neural Network (GNN) models. The primary objective of XAI for Graphs is to foster a sense of trust and enable effective utilization of these models by human users~\cite{XAI4GNN_survey}. Motivated by previous studies on the explainability of canonical graphs~\cite{GNNExplainer}, we propose to enhance the explainability of STGNN by searching for subgraph $\mathcal{G}_S$ based on the STG $\mathcal{G}$ and the ground-truth label $\mathbf{Y}$:
\begin{align}
    \mathcal{G}_S = \mathop{\arg\max}_{\mathcal{G}_S} I(\mathbf{Y}, \mathcal{G}_S) &= H(\mathbf{Y}) - H(\mathbf{Y}|\mathcal{G}_S) \label{eq:xai4g}
\end{align}
where $I(\cdot)$ denotes the mutual information function, $H(\cdot)$ represents the information entropy, $\mathcal{G}_S=(\mathcal{V}_S, \mathcal{E}_S, \mathbf{A}_S, \mathbf{X}_S)$ represents the subgraph of $\mathcal{G}$ with the sub-node set $\mathcal{V}_S$, the sub-edge set $\mathcal{E}_S$, the sub-adjacency matrix $A_S$ and the sub-feature matrix $\mathbf{X}_S$. The model is optimized to find subgraph $\mathcal{G}_S$ which makes the most prominent contribution to predictions made by model $f$, which helps humans comprehend the black-box GNN model $f$ intuitively. 
Next, we introduce the definition of two categories of explainability approaches on graphs, \ie, \emph{post-hoc} and \emph{intrinsic}, as follows. \\\vspace{-0.12in}

\noindent (i) \textbf{Post-hoc}. With the GNN model $f$, the \emph{post-hoc} model aims to learn an explainability function $\Gamma$ to identify the subgraph $\mathcal{G}_S$ contributing to the performance of $f$ the most:
\begin{align}
    \mathcal{G}_S = \Gamma(\mathcal{G}, \mathbf{Y}, f) ~~~~
    \mathrm{ s.t. } \max_{\mathcal{G}_S} I(\mathbf{Y}, \mathcal{G}_S)
\end{align}

\noindent (ii) \textbf{Intrinsic}. Distinct from \emph{post-hoc} methods, the goal of \textit{intrinsic} approaches is to learn a unified model $f^{\prime}$ to simultaneously predict the target graph signals and identify the subgraph $\mathcal{G}_S$ that impacts the model $f^{\prime}$ the most, which is defined as below:
\begin{align}
    \hat{\mathbf{Y}}, \mathcal{G}_S = f^{\prime}(\mathcal{G}) ~~~~
    \mathrm{ s.t. } \min \mathcal{L}(\mathbf{Y}, \hat{\mathbf{Y}}) 
    \wedge \max_{\mathcal{G}_S} I(\mathbf{Y}, \mathcal{G}_S)
\end{align}
where $\mathcal{L}$ denotes a specific loss function supervising the predictive task on graphs, and $\hat{\mathbf{Y}}$ presents the predicted results of the model.
We summarize the notations frequently used in our paper in Table~\ref{tab:notation}. 

\begin{table}
\centering
\caption{Description of notations in our paper.}\label{tab:notation}
\vspace{-0.1in}
\resizebox{.5\textwidth}{!}{\begin{tabular}{c|c} 
\hline
\textbf{Notations} & \textbf{Description}  \\ 
\hline
$\mathbf{X}\in \mathbb{R}^{T\times N\times F}$               & Original STG feature matrix.                                 \\
$\mathbf{X}^{(0)}\in\mathbb{R}^{T\times N\times d}$          & Initialized STG embeddings.                                  \\
$\mathbf{X}^{(s)} = \{\mathbf{x}_j^{(s)} \in\mathbb{R}^{d_s}, 1\leq j\leq N\}$ & Spatial feature matrix.                                      \\
$\mathbf{H}^{(s)} = \{\vec{h}_j^{(s)}\in\mathbb{R}^{d_s}, 1\leq j\leq N\}$ & Extracted spatial embeddings.                                \\
$\mathbf{H}^{\prime(s)}\in\mathbb{R}^{T\times N\times d}$    & Intermediate feature matrix in the proposed STG encoder.     \\
$\mathbf{X}^{(t)}= \{\vec{x}_i^{(t)}\in \mathbb{R}^{d_{t}}, 1\leq i \leq T\}$ & Temporal feature matrix.                                     \\
$\mathbf{H}^{(t)} = \{\vec{h}_i^{(t)}\in \mathbb{R}^{d_{t}}, 1\leq i \leq T\}$ & Extracted temporal embeddings.                               \\
$\mathbf{H} = \mathbf{H}^{\prime (t)}\in \mathbb{R}^{T\times N\times d}$ & The final output feature matrix of the proposed STG encoder. \\
$\mathcal{G}^{(s)}(\mathcal{V}^{(s)}, \mathcal{E}^{(s)}, A^{(s)}, \mathbf{X}^{(s)})$ & Spatial graph with spatial adjacency matrix $A^{(s)}\in \mathbb{R}^{N\times N}$. \\
$\mathcal{G}^{(t)}(\mathcal{V}^{(t)}, \mathcal{E}^{(t)}, A^{(t)}, \mathbf{X}^{(t)})$ & Temporal graph with temporal adjacency matrix $A^{(t)}\in \mathbb{R}^{T\times T}$. \\
$E^{(s)}\in \mathbb{R}^{N\times D}$                          & Learnable spatial position.                                  \\
$\mathbf{E}^{(ToD)}\in\mathbb{R}^{T\times d}$                & Learnable *time of day* embeddings                           \\
$\mathbf{E}^{DoW}\in\mathbb{R}^{T\times d}$                  & Learnable *day of week* embeddings                           \\
$\mathcal{G}_S=(\mathcal{V}_S, \mathcal{E}_S, A_S, \mathbf{X})$ & Explainable subgraph.                                        \\
$\mathbb{P}(\mathcal{G}_S|\mathcal{G})$                      & Variational approximation of explainable subgraph given the original graph. \\
$\mathbb{Q}_1(\mathbf{Y}|\mathcal{G}_S)$                     & Variational approximation of prediction given the explainable subgraph. \\
$\mathbb{Q}_2(\mathcal{G}_S)$                                & The priori distribution of the explainable subgraph.         \\
\hline
\end{tabular}}
\vspace{-0.1in}
\end{table}
\section{Methodology}
\label{sec:solution}
\begin{figure*}
    \centering
    \includegraphics[width=0.98\textwidth]{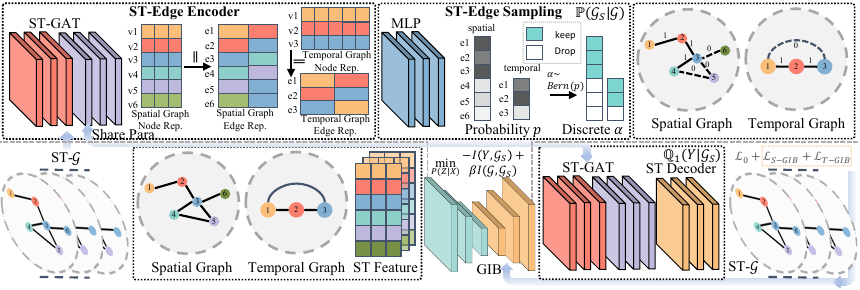}
    \vspace{-0.15in}
    \caption{The overall framework of the proposed \model\ framework is as follows: the STG is decoupled into spatial and temporal graph structures to capture spatio-temporal features. These structures are then fed into the structure-distilled GIB module with the ST-Edge Encoder, resulting in spatial and temporal graph edge representations. Additionally, ST-Edge Sampling is employed to obtain explainable spatial and temporal graph structures. Finally, ST-GAT is utilized to encode spatial and temporal dependencies on the explainable structures, ultimately producing the final results.}
    \vspace{-0.15in}
    \label{fig:overall}
\end{figure*}

In this section, we provide a detailed description of the technical aspects and theoretical analysis of our \model\ framework. Our framework encompasses a unified STGNN encoder that employs spatio-temporal graph attention networks to reason about spatio-temporal dependencies. Additionally, we propose the structure-distilled Graph Information Bottleneck (GIB) for STG to select explainable subgraph structures benefiting the downstream forecasting. The overall architecture of \model\ is illustrated in Figure~\ref{fig:overall}.


\vspace{-0.1in}
\subsection{Spatio-Temporal Graph Attention Networks}
\label{sec:stgat}
\subsubsection{\bf Spatial Relation Learning}
Inspired by GNN's strength of reasoning the complicated correlations~\cite{graphAE, GIN, GAT}, especially in spatio-temporal modeling~\cite{DCRNN, STGCN},
we propose a unified GNN encoder which adapts graph attention networks~\cite{GAT} to capture the spatio-temporal dependencies. Following~\cite{STFGNN}, we could employ a unified GNN-based framework to capture spatio-temporal dependencies on a unified spatio-temporal graph structure $\mathbf{A}\in \mathbb{R}^{TN\times TN}$. 
To avoid the enormous time complexity of STG learning, we decouple the joint graph into a temporal graph and a spatial graph.
Primarily, STG feature matrix $\mathbf{X}\in \mathbb{R}^{T\times N\times F}$ is embeded into a $d$-dimensional latent space with the fully connected layer:

\begin{align}
    \mathbf{X}^{(0)} &= \mathbf{X} \cdot \mathbf{W}^{(0)} + \mathbf{b}^{(0)} \label{eq:start_fc}
\end{align}
where $\mathbf{X}^{(0)}\in\mathbb{R}^{T\times N\times d}$ represents initial embeddings of the STG. $\mathbf{W}^{(0)}\in \mathbb{R}^{F\times d}$, $\mathbf{b}^{(0)}\in \mathbb{R}^{d}$ denote the weight and bias matrices. Furthermore, to individually encode spatial and temporal dynamics with our GAT, $\mathbf{X}^{(0)}$ is converted to spatial embeddings $\mathbf{X}^{(s)} = \{\mathbf{x}_j^{(s)} \in\mathbb{R}^{d_s}, 1\leq j\leq N\}$ employing linear transformation by:

\begin{align}
    \label{eq:fcs}
    \mathbf{x}_j^{(s)} = \sum_{i=1}^T \mathbf{X}^{(0)}_{i,j,:}\mathbf{W}_i^{(s)} + \mathbf{b}^{(s)}
\end{align}
where $\mathbf{W}^{(s)}\in \mathbb{R}^{T\times d\times d_{s}}$ and $\mathbf{b}^{(s)}\in \mathbb{R}^{d_{s}}$ indicate weight and bias parameters.
In this stage, we utilize the spatial subgraph in the STG $\mathcal{G}$, which is defined by $\mathcal{G}^{(s)}=(\mathcal{V}^{(s)}, \mathcal{E}^{(s)}, \mathbf{A}^{(s)}, \mathbf{X}^{(s)})$,
where $\mathbf{A}^{(s)}\in \mathbb{R}^{N\times N}$ denotes the spatial adjacency matrix recording the spatial node-wise correlations. 
Regarding the spatial graph reasoning, we employ GAT with stacked multi-head graph attention layers, where the $K$-head graph attention layer is defined as below:
\begin{align}
    &\mathbf{h}^{(s)}_j = \sum_{k=1}^K\sum_{j\prime\in \mathcal{N}(j) \cup \{j\}} \alpha_{j,j'}^{k} \cdot \mathbf{W}^{k} \mathbf{x}^{(s)}_{j'} \nonumber \\ 
    \alpha_{j,j'} = &\frac{\exp(\sigma(\vec{a}^\top[\mathbf{W} \Vec{x}^{(s)}_{j} + \mathbf{W} \Vec{x}^{(s)}_{j'}]))}{\sum_{j'\in \mathcal{N}(j) \cup \{j\}}\exp(\sigma(\vec{a}^\top[\mathbf{W} \Vec{x}^{(s)}_j + \mathbf{W} \Vec{x}^{(s)}_{j'}]))} \label{eq:gat}
\end{align}
where $\mathcal{N}(j)$ represents the set of neighbors of the $j$-th region according to $\mathbf{A}^{(s)}$, $\vec{a}\in \mathbb{R}^{d_{s}}$ represents the weight vector, $\mathbf{W}\in \mathbb{R}^{d_{s} \times d_{s}}$ indicates the weight parameters, and $\sigma(\cdot)$ denotes the LeakyReLU activation function. 
With multiple GAT layers, we gain the extracted spatial embeddings $\mathbf{H}^{(s)} = \{\Vec{h}_j^{(s)}\in\mathbb{R}^{d_s}, 1\leq j\leq N\}$. 
Then we transform $\mathbf{H}^{(s)}$ into the spatio-temporal embedding space to get $\mathbf{H}^{\prime(s)}\in\mathbb{R}^{T\times N\times d}$ utilizing fully-connected layer with weight matrix $\mathbf{W}^{(1)}\in \mathbb{R}^{T\times d \times d_{s}}$ and bias parameters $\mathbf{B}^{(1)}\in\mathbb{R}^{T\times d}$ as:
\begin{align}
    \label{eq:final}
    \mathbf{H}^{\prime(s)}_{i,j,:} = \mathbf{W}_{i}^{(1)}\cdot \mathbf{h}_j^{(s)} + \mathbf{B}^{(1)}_i
\end{align}

\subsubsection{\bf Temporal Relation Learning}
We follow the similar relation learning paradigm to model the temporal graph $\mathcal{G}^{(t)}=(\mathcal{V}^{(t)}, \mathcal{E}^{(t)}, \mathbf{A}^{(t)}, \mathbf{X}^{(t)})$, where $\mathbf{A}^{(t)}\in \mathbb{R}^{T\times T}$ indicates the temporal adjacency matrix revealing the correlations among time steps, and $\mathbf{X}^{(t)}= \{\Vec{x}_i^{(t)}\in \mathbb{R}^{d_{t}}, 1\leq i\leq T\}$ represents the temporal feature matrix. $\mathbf{X}^{(t)}$ is transformed from $\mathbf{H}^{\prime (s)}$ by utilizing the similar fully connected layer as Eq~\ref{eq:fcs}.
To model the temporal dynamics, stacked multi-head GAT layers defined analogously as Eq~\ref{eq:gat} are utilized to generate the temporal feature matrix $\mathbf{H}^{(t)} = \{\Vec{h}_i^{(t)}\in \mathbb{R}^{d_{t}}, 1\leq i\leq T\}$. 
Eventually, we transform the temporal features $\mathbf{H}^{(t)}$ into the final spatio-temporal embedding matrix $\mathbf{H} = \mathbf{H}^{\prime (t)}\in \mathbb{R}^{T\times N\times d}$ adopting the similar transformation function as Eq~\ref{eq:final}. 
So, we summarize how to construct spatial and temporal graphs as follows: 
\noindent(i) \textbf{spatial graph ($A^{(s)}$):} 
Spatial graph represents the correlations between spatial units. For the two common types of spatio-temporal prediction, \ie, graph-based and grid-based~\cite{DL-Traff}, we can construct graphs using a thresholded Gaussian kernel~\cite{DCRNN} and considering neighboring regions as neighbors~\cite{ST-HSL, ST-SHN}, respectively.
\noindent(ii) \textbf{temporal graph ($A^{(t)}$):} 
Temporal graph represents the correlations between temporal representations at different time steps.
Formally, if the historical time step is $T$, we have temporal graph $A^{(t)}\in \mathbb{R}^{T\times T}$ and $A^{(t)}_{i,j} = 1$ for arbitrary $i, j$. This means that we assume that every time step influence others originally. 
Applying GAT for message passing on the temporal graph is equivalent to existing works~\cite{GMAN} that utilize self-attention to capture temporal correlations.

\subsubsection{\bf Position-Aware STG Prediction}
To enhance the modeling of spatio-temporal contexts in the model inference phase of our \model, we propose to inject spatial and temporal positional embeddings into the foregoing STG relational embeddings $\mathbf{H}$. In specific, multiple free-form embeddings are leveraged by our \model: the region representations $\mathbf{E}^{(s)}\in\mathbb{R}^{N\times d}$, the \textit{time of day} embeddings $\mathbf{E}^{(ToD)}\in\mathbb{R}^{T\times d}$, and the \textit{day of week} embeddings $\mathbf{E}^{DoW}\in\mathbb{R}^{T\times d}$. 
For implementation, we randomly initialize a tensor $E^{(s)}\in \mathbb{R}^{N\times D}$, and the value of the tensor could be updated during back propagation (i.e., learnable). 
As to temporal positional embeddings, we randomly initialize a \emph{time of day} tensor $E^{({ToD})}_{\text{all}}\in \mathbb{R}^{288\times D}$ and a \emph{day of week} tensor $E^{({DoW})}_{\text{all}}\in \mathbb{R}^{7\times D}$, where 288 denotes a day has 288 time steps (for 5 min interval) and 7 denotes a week has 7 days. The input \emph{time of day} and \emph{day of week} index of the STG query \emph{time of day} and \emph{day of week} tensors to obtain temporal positional embeddings.
Then, \model\ makes predictions as follows:
\begin{align}
    \mathbf{Y} = \mathbf{MLP}_1(\mathbf{H} \| \mathbf{E}^{(s)} \| \mathbf{E}^{(ToD)} \| \mathbf{E}^{(DoW)} \| \mathbf{MLP}_2(\mathbf{X})) \label{eq:pred_layer}
\end{align}
where $\|$ denotes concatenation, $\mathbf{MLP}_1(\cdot)$ and $\mathbf{MLP}_2(\cdot)$ denote two multi-layer perceptrons for making final predictions and leveraging low-level features $\mathbf{X}$, respectively. $\mathbf{Y}$ denotes the predictions for future STG attributes using the position-aware STG embeddings.

\vspace{-0.05in}
\subsection{Spatio-Temporal Explainability with GIB}
\subsubsection{\bf GIB-based Explainable Structure Distillation}
The Graph Information Bottleneck (GIB) technique is designed to compress graph-structured data into low-dimensional representations that exhibit strong correlation with downstream labels. These compressed representations capture a subset of the original information while effectively accounting for the labels in subsequent tasks. As a result, GIB has gained recognition as an explainable model in certain literature, such as \cite{GSAT, Trustai}. The underlying principle of GIB is to optimize the embeddings by minimizing the following objective:
\begin{align}
    \min_{\mathbb{P}(\mathbf{Z}_X|\mathcal{G})} -I(Y,\mathbf{Z}_X)+\beta I(\mathcal{G}, \mathbf{Z}_X) \label{eq:gib}
\end{align}
The hidden representations of the graph feature matrix $\mathbf{X}$ are denoted as $\mathbf{Z}_X$. While the conventional GIB generates low-dimensional representations that capture the reasoning behind downstream labels, these dense hidden embeddings are often challenging for humans to comprehend. This limitation significantly restricts the applicability of using the conventional GIB for model interpretation. In order to address the objective of developing explainable spatio-temporal graph (STG) models, as outlined in Eq~\ref{eq:xai4g}, we draw inspiration from~\cite{GSAT} and propose the structure-distilled GIB approach. This approach applies the Information Bottleneck (IB) principle to distilled subgraph structures, enabling the acquisition of a small subset of interpretable STG structures. Specifically, the objective of our structure-distilled GIB is defined as follows:
\begin{align}
    \min_{\mathbb{P}(\mathcal{G}_S|\mathcal{G})} -I(\mathbf{Y},\mathcal{G}_S)+\beta \cdot I(\mathcal{G}, \mathcal{G}_S) \label{eq:xgib_1}
\end{align}
\noindent The subgraph $\mathcal{G}_S = (\mathcal{V}_S, \mathcal{E}_S, A_S, \mathbf{X}_S)$ represents the distilled subgraph obtained from the conditional probability distribution given the original graph $\mathcal{G}$. In real-world scenarios, the graph structures play a crucial role in spatio-temporal graphs and are easier for humans to interpret as a rationale for model inference. Therefore, we prioritize the use of subgraph structures for interpretation purposes and simplify the objective presented in Equation~\ref{eq:xgib_1} by defining the subgraph as $\mathcal{G}_S = (\mathcal{V}_S, \mathcal{E}_S, A_S, \mathbf{X})$.

\subsubsection{\bf Variational Bounds for Structure-Distilled GIB}
Since the mutual information terms $I(\mathbf{Y},\mathcal{G}_S)$ and $I(\mathcal{G}, \mathcal{G}_S)$ are intractable, we resort to using variational bounds to estimate each term in the objective. For the lower bound of the first term $I(\mathbf{Y},\mathcal{G}_S)$, we can utilize the fact that $\text{KL}[\mathbb{P}(\mathbf{Y}|\mathcal{G}_S), \mathbb{Q}_1(\mathbf{Y}|\mathcal{G}_S)] \geq 0$, where $\mathbb{Q}_1(\mathbf{Y}|\mathcal{G}_S)$ represents an arbitrary distribution of $\mathbf{Y}$ given $\mathcal{G}_S$. Thus, we obtain:
\begin{align}
    I(\mathbf{Y},\mathcal{G}_S) &= \mathbb{E}_{\mathbf{Y}, \mathcal{G}_S}[\log\frac{\mathbb{P}(\mathbf{Y}|\mathcal{G}_S)}{\mathbb{P}(\mathbf{Y})}]
     \geq \mathbb{E}_{\mathbf{Y}, \mathcal{G}_S} [\log\mathbb{Q}_1(\mathbf{Y}|\mathcal{G}_S)] \label{eq:lower}
\end{align}
The expression $\log\mathbb{Q}_1(\mathbf{Y}|\mathcal{G}_S)$ also represents the variational approximation of $\mathbb{P}(\mathbf{Y}|\mathcal{G}_S)$, which can be modeled using neural networks within an end-to-end framework. Specifically, $\log\mathbb{Q}_1(\mathbf{Y}|\mathcal{G}_S)$ aims to predict the results based on the subgraph $\mathcal{G}_S$. Regarding the upper bound of the second term $I(\mathcal{G}, \mathcal{G}_S)$, we can establish that $\text{KL}[\mathbb{P}(\mathcal{G}_S), \mathbb{Q}_2(\mathcal{G}_S)] \geq 0$ holds true. We can formalize it as follows:
\begin{align}
    I(\mathcal{G}, \mathcal{G}_S) = \mathbb{E}_{\mathcal{G}, \mathcal{G}_S}[\log\frac{\mathbb{P}(\mathcal{G}_S|\mathcal{G})}{\mathbb{P}(\mathcal{G}_S)}] 
    \leq \mathbb{E}_{\mathcal{G}} [\text{KL}(\mathbb{P}(\mathcal{G}_S|\mathcal{G})\| \mathbb{Q}_2(\mathcal{G}_S))] \label{eq:upper}
\end{align}
$\mathbb{Q}_2(\mathcal{G}_S)$ is the variational approximation for the marginal distribution $\mathbb{P}(\mathcal{G}_S)$. The ultimate objective for Eq~\ref{eq:xgib_1} is defined as:
\begin{align}
    \min_{\mathbb{P}(\mathcal{G}_S|\mathcal{G})} -\mathbb{E}_{\mathbf{Y}, \mathcal{G}_S} [\log\mathbb{Q}_1(\mathbf{Y}|\mathcal{G}_S)]+\beta \mathbb{E}_{\mathcal{G}} [\text{KL}(\mathbb{P}(\mathcal{G}_S|\mathcal{G})\| \mathbb{Q}_2(\mathcal{G}_S))] \label{eq:xgib_2}
\end{align}

\subsubsection{\bf Spatio-Temporal GIB Characterization}
To minimize the upper bound in Eq~\ref{eq:xgib_2} for our structure-distilled GIB, it is necessary to characterize the distributions $\mathbb{P}(\mathcal{G}_S|\mathcal{G})$, $\mathbb{Q}_1(\mathbf{Y}|\mathcal{G}_S)$ and $\mathbb{Q}_2(\mathcal{G}_S)$.

\textbf{i)} $\mathbb{P}(\mathcal{G}_S|\mathcal{G})$: 
To extract the influential subgraph $\mathcal{G}_S$ from the original graph $\mathcal{G}$, we incorporate randomness into the instantiated networks.
In particular, we begin by embedding the spatio-temporal graphs $\mathcal{G}^{(s)}$ and $\mathcal{G}^{(t)}$ using a unified STGNN encoder. 
This process yields the spatio-temporal node representations $\mathbf{H}^{(s)} = \{ \Vec{h}_j^{(s)}\in \mathbb{R}^{d_{s}}\}$ and $\mathbf{H}^{(t)} = \{\Vec{h}_i^{(t)}\in \mathbb{R}^{d_{t}}\}$. Next, we employ the concatenation operator $|$ and an MLP $\mathcal{F}_{\Theta}$ with parameters $\Theta$ to encode the spatio-temporal edge representation. This encoding step is defined as:
\begin{align}
    \vec{h}^{(s)}_{vu} &= \mathcal{F}_{\Theta^{(s)}}(\vec{h}^{(s)}_{v}\| \vec{h}^{(s)}_{u}), \mathrm{s.t.}, u\in \mathcal{N}(v) \nonumber \\ 
    \vec{h}^{(t)}_{vu} &= \mathcal{F}_{\Theta^{(t)}}(\vec{h}^{(t)}_{v}\| \vec{h}^{(t)}_{u}), \mathrm{s.t.}, u\in \mathcal{N}(v) \label{eq:edge_concat}
\end{align}
where $\mathcal{N}(v)$ denotes the neighbor set of node $v$. Subsequently, we employ the Gumbel-Softmax reparameterization trick~\cite{softmax_1,softmax_2} to compute the spatio-temporal probabilities ${p}^{(s)}_{vu}$ and ${p}^{(t)}_{vu}$ for each edge in a differentiable manner. This enables us to have:
\begin{align}
    {p}^{(s)}_{vu} = \sigma((\vec{h}^{(s)}_{vu}+\textsl{g})/\tau),~~~~ 
    {p}^{(t)}_{vu} = \sigma((\vec{h}^{(t)}_{vu}+\textsl{g})/\tau) \label{eq:sampling}
\end{align}
where $\textsl{g}$ is a set of i.i.d. samples drawn from a Gumbel(0,1) distribution, and $\tau$ is the temperature parameter that controls the smoothness of the resulting distribution. Consequently, we obtain the spatio-temporal explainable subgraph structures $\mathbf{A}^{(s)}_S$ and $\mathbf{A}^{(t)}_S$:
\begin{align}
    \mathbf{A}^{(s)}_S = \alpha^{(s)}\odot A^{(s)},~~~~
    \alpha^{(s)}_{vu} \sim \text{Bern}({p}^{(s)}_{vu}) \nonumber \\ 
    \mathbf{A}^{(t)}_S = \alpha^{(t)}\odot A^{(t)},~~~~
    \alpha^{(t)}_{vu} \sim \text{Bern}({p}^{(t)}_{vu}) \label{eq:xsubgraph}
\end{align}
The symbol $\odot$ is the element-wise product. $\alpha^{(s)}$ and $\alpha^{(t)}$ are the spatio-temporal subgraph selectors used to extract the explainable subgraphs. Consequently, the instantiation of the spatio-temporal term $\mathbb{P}(\mathcal{G}_S|\mathcal{G})$ is as follows:
\begin{align}
    \mathbb{P}(\mathcal{G}_S^{(s)}|\mathcal{G}^{(s)}) &= \prod_{v,u\in \mathcal{V}^{(s)}} \mathbb{P}(\alpha^{(s)}_{vu}|{p}^{(s)}_{vu}) \nonumber \\ 
    \mathbb{P}(\mathcal{G}_S^{(t)}|\mathcal{G}^{(t)}) &= \prod_{v,u\in \mathcal{V}^{(t)}} \mathbb{P}(\alpha^{(t)}_{vu}|{p}^{(t)}_{vu})
\end{align}

\textbf{ii)} $\mathbb{Q}_1(\mathbf{Y}|\mathcal{G}_S)$: 
The goal of this variational approximation is to infer the spatio-temporal dynamics based solely on the extracted spatio-temporal explainable subgraphs. To achieve this, we utilize the proposed spatio-temporal graph attention network (ST-GAT) architecture, which consists of the same set of learnable parameters as introduced in Section~\ref{sec:stgat}. It is important to note that when calculating $\mathbb{Q}_1(\mathbf{Y}|\mathcal{G}_S)$, our ST-GAT performs message propagation exclusively along the sampled explainable edges and nodes.

\textbf{iii)} $\mathbb{Q}_2(\mathcal{G}_S)$: 
Regarding the prior distribution $\mathbb{Q}_2(\mathcal{G}_S)$, we have the following formalizations for the spatial and temporal graphs: 
\begin{align}
    \mathbb{Q}_2(\mathcal{G}_S^{(s)}) &= \sum_{\mathcal{G}^{(s)}} \mathbb{P}(\mathcal{G}^{(s)}, \mathcal{G}_S^{(s)}) = \sum_{\mathcal{G}^{(s)}} \mathbb{P}(\mathcal{G}_S^{(s)}|\mathcal{G}^{(s)})\mathbb{P}(\mathcal{G}^{(s)}) \nonumber \\ 
    \mathbb{Q}_2(\mathcal{G}_S^{(t)}) &= \sum_{\mathcal{G}^{(t)}} \mathbb{P}(\mathcal{G}^{(t)}, \mathcal{G}_S^{(t)}) = \sum_{\mathcal{G}^{(t)}} \mathbb{P}(\mathcal{G}_S^{(t)}|\mathcal{G}^{(t)})\mathbb{P}(\mathcal{G}^{(t)}) 
\end{align}
Following~\cite{GSAT}, for the given spatio-temporal graphs $\mathcal{G}^{(s)}$ with $n^{(s)}$ edges and $\mathcal{G}^{(t)}$ with $n^{(t)}$ edges, we sample prior spatio-temporal selectors $\alpha^{\prime(s)}$ and $\alpha^{\prime(t)}$, which is defined as below:
\begin{align}
    &\alpha^{\prime(s)}\sim \text{Bern}(r^{(s)}), \alpha^{\prime(t)} \sim \text{Bern}(r^{(t)})\nonumber \\ 
    &\mathbb{Q}_2(\mathcal{G}_S^{(s)}) =  \sum_{n} \mathbb{P}(\alpha^{\prime(s)}|n^{(s)})\mathbb{P}(n^{(s)}) \nonumber \\ 
    &\mathbb{Q}_2(\mathcal{G}_S^{(t)}) =  \sum_{n} \mathbb{P}(\alpha^{\prime(t)}|n^{(t)})\mathbb{P}(n^{(t)})
\end{align}
The selector $\alpha^{\prime}_{vu} = 1$ indicates that the edge $(v,u) \in \mathcal{E}$ in graph $\mathcal{G}$. The hyperparameters $r^{(s)}$ and $r^{(t)}$ are used for sampling. Since $\mathbb{P}(n^{(s)})$ and $\mathbb{P}(n^{(t)})$ are constants and independent of $\alpha^{\prime(s)}$ and $\alpha^{\prime(t)}$, we can simplify the expression, and ultimately we obtain:
\begin{align}
    \mathbb{Q}_2(\mathcal{G}_S^{(s)}) & = \mathbb{P}(n^{(s)}) \prod_{v,u = 1}^{n} \mathbb{P}(\alpha^{\prime(s)}_{vu}) \nonumber \\ 
    \mathbb{Q}_2(\mathcal{G}_S^{(t)}) & = \mathbb{P}(n^{(t)}) \prod_{v,u = 1}^{n} \mathbb{P}(\alpha^{\prime(t)}_{vu})
\end{align}

\vspace{-0.1in}
\subsection{Model Optimization}
In our \model\ framework, we optimize towards the objective of structure-distilled GIB as defined in Equation~\ref{eq:xgib_2}. To infer the downstream labels $\textbf{Y}$ using the explainable subgraph $\mathcal{G}_S$, we utilize different loss functions depending on the specific spatio-temporal prediction tasks.
For instance, when predicting future traffic volumes, we employ the Huber loss~\cite{huber1992robust}.
\begin{equation}
    \label{eq6}
    \mathcal{L}_0(\mathbf{Y}, \hat{\mathbf{Y}}) = \mathcal{H}(\mathbf{Y}, \hat{\mathbf{Y}})=\left\{
    \begin{aligned}
    &\frac{1}{2}(\mathbf{Y} - \hat{\mathbf{Y}})  , & \left\lvert \mathbf{Y} - \hat{\mathbf{Y}} \right\rvert  \leq  \delta  \\
    &\delta(\left\lvert \mathbf{Y} - \hat{\mathbf{Y}} \right\rvert - \frac{1}{2}\delta)  , & otherwise
    \end{aligned}
    \right.
\end{equation}
where $\delta$ denotes the hyperparameter for threshold. For the crime prediction, we instead utilize the mean absolute error (MSE) loss following~\cite{ST-HSL} and have the following loss: $\mathcal{L}_0(\mathbf{Y}, \hat{\mathbf{Y}}) = \left\lVert \mathbf{Y} - \hat{\mathbf{Y}} \right\rVert^2_2$.
For the second item in the upper-bound GIB objective (Eq~\ref{eq:xgib_2}), we employ specific loss functions for the spatial and temporal explainable subgraphs, respectively.
\begin{align}
    \mathcal{L}_{\text{S-GIB}} &=  \mathbb{E}_{\mathcal{G}^{(s)}} [\text{KL}(\mathbb{P}(\mathcal{G}_S^{(s)}|\mathcal{G}^{(s)})\| \mathbb{Q}_2(\mathcal{G}_S^{(s)}))] \nonumber \\ 
    &= \sum_{(v, u)\in \mathcal{E}^{(s)}}p_{vu}^{(s)}\log\frac{p_{vu}^{(s)}}{r^{(s)}} + (1 - p_{vu}^{(s)}) \log\frac{1 - p_{vu}^{(s)}}{1 - r^{(s)}} + C \nonumber \\
    \mathcal{L}_{\text{T-GIB}} &=  \mathbb{E}_{\mathcal{G}^{(t)}} [\text{KL}(\mathbb{P}(\mathcal{G}_S^{(t)}|\mathcal{G}^{(t)})\| \mathbb{Q}_2(\mathcal{G}^{(t)}_S))] \nonumber \\ 
    &= \sum_{(v, u)\in \mathcal{E}^{(t)}}p_{vu}^{(t)}\log\frac{p_{vu}^{(t)}}{r^{(t)}} + (1 - p_{vu}^{(t)}) \log\frac{1 - p_{vu}^{(t)}}{1 - r^{(t)}} + C \label{tb:ib_loss}
\end{align}
Combining the above loss functions, the optimization for our \model\ framework is to minimize the below jointly-training objective, with weighing hyperparameters $\lambda_1, \lambda_2$. 
\begin{align}
    \mathcal{L} = \mathcal{L}_{0} + \lambda_1\mathcal{L}_{\text{S-GIB}} + \lambda_2\mathcal{L}_{\text{T-GIB}} \label{eq:loss_final}
\end{align}

\section{Experiments}
\label{sec:exp}
\begin{table*}[t]
  \vspace{-0.15in}
  \centering
  \caption{Performance comparison of different methods on PEMS4, 7, 8 datasets.}
  \vspace{-0.1in}
  \footnotesize
  \setlength{\tabcolsep}{0.6mm}
  {
  \begin{tabular}{c|c|ccccccccccccccccc|cc} 
  \hline
  \multicolumn{2}{c|}{Model}        & {HA} & {VAR} & {DCRNN} & {STGCN}  & {GWN} & {\begin{tabular}[c]{@{}c@{}}AST\\GCN\end{tabular}}  & {\begin{tabular}[c]{@{}c@{}}STS\\GCN\end{tabular}} & {\begin{tabular}[c]{@{}c@{}}Stem\\GNN\end{tabular}} & {AGCRN} & {STFGNN} & {STGODE} & {\begin{tabular}[c]{@{}c@{}}Z-\\ GCNETs\end{tabular}} & {\begin{tabular}[c]{@{}c@{}}TAMP-\\ S2GCNets\end{tabular}} & {GMSDR} & {FOGS} & {\begin{tabular}[c]{@{}c@{}}STG-\\ NCDE\end{tabular}} & {\begin{tabular}[c]{@{}c@{}}DSTA\\GNN\end{tabular}} & {\begin{tabular}[c]{@{}c@{}}\model-\\ CGIB\end{tabular}} & {{\model}}  \\ 
  \hline
  \multirow{3}{*}{\rotatebox{90}{PEMS4}} & MAE     & 38.03               & 24.54                & 21.22                  & 21.16                   & 24.89                & 22.93                    & 21.19                   & 21.61                    & 19.83                  & 19.83                   & 20.84                   & 19.50                     & 19.74                                                                     & 20.49                  & 19.74                 & 19.21                     & 19.30                    & 19.14                       & \textbf{18.57}           \\
                          & RMSE    & 59.24               & 38.61                & 33.44                  & 34.89                   & 39.66                & 35.22                    & 33.65                   & 33.80                    & 32.26                  & 31.88                   & 32.82                   & 31.61                     & 31.74                                                                     & 32.13                  & 31.66                 & 31.09                     & 31.46                    & 30.77                       & \textbf{30.14}           \\
                          & MAPE(\%)    & 27.88             & 17.24              & 14.17                & 13.83                 & 17.29              & 16.56                  & 13.90                 & 16.10                  & 12.97                & 13.02                 & 13.77                 & 12.78                   & 13.22                                                                   & 14.15                & 13.05               & 12.76                   & 12.70                  & 12.91                     & \textbf{12.13}         \\ 
  \hline
  \multirow{3}{*}{\rotatebox{90}{PEMS7}} & MAE     & 45.12               & 50.22                & 25.22                  & 25.33                   & 26.39                & 24.01                    & 24.26                   & 22.23                    & 22.37                  & 22.07                   & 22.99                   & 21.77                     & 21.84                                                                     & 22.27                  & 21.28                 & 20.53                     & 21.42                    & 20.55                       & \textbf{20.00}           \\
                          & RMSE    & 65.64               & 75.63                & 38.61                  & 39.34                   & 41.50                & 37.87                    & 39.03                   & 36.46                    & 36.55                  & 35.80                   & 37.54                   & 35.17                     & 35.42                                                                     & 34.94                  & 34.88                 & 33.84                     & 34.51                    & 35.12                       & \textbf{33.45}           \\
                          & MAPE(\%)    & 24.51             & 32.22              & 11.82                & 11.21                 & 11.97              & 10.73                  & 10.21                 & 9.20                   & 9.12                 & 9.21                  & 10.14                 & 9.25                    & 9.24                                                                    & 9.86                 & 8.95                & 8.80                    & 9.01                   & 8.61                      & \textbf{8.51}          \\ 
  \hline
  \multirow{3}{*}{\rotatebox{90}{PEMS8}} & MAE     & 34.86               & 19.19                & 16.82                  & 17.50                   & 18.28                & 18.25                    & 17.13                   & 15.91                    & 15.95                  & 16.64                   & 16.81                   & 15.76                     & 16.36                                                                     & 16.36                  & 15.73                 & 15.45                     & 15.67                    & 14.87                       & \textbf{14.59}           \\
                          & RMSE    & 52.04               & 29.81                & 26.36                  & 27.09                   & 30.05                & 28.06                    & 26.80                   & 25.44                    & 25.22                  & 26.22                   & 25.97                   & 25.11                     & 25.98                                                                     & 25.58                  & 24.92                 & 24.81                     & 24.77                    & 24.07                       & \textbf{23.91}           \\
                          & MAPE(\%)    & 24.07             & 13.10              & 10.92                & 11.29                 & 12.15              & 11.64                  & 10.96                 & 10.90                  & 10.09                & 10.60                 & 10.62                 & 10.01                   & 10.15                                                                   & 10.28                & 9.88                & 9.92                    & 9.94                   & 10.26                     & \textbf{9.80}          \\
  \hline
  \end{tabular}}\label{tb:cmp1}
  \end{table*}
  
To evaluate the performance of \model\ in terms of predictive accuracy and explainability, we conduct extensive experiments on three real-world traffic datasets and two crime datasets by answering questions: 
\textbf{RQ1}: How does \model\ perform while predicting future traffic volume and crimes compared to various state-of-the-art baselines?
\textbf{RQ2}: How does the \model\ framework compare to different state-of-the-art explainable models in terms of quantitative explainability?
\textbf{RQ3}: How do key components contribute to the performance of \model\ framework?
\textbf{RQ4}: How does the \model\ framework perform in terms of generalization and robustness?
\textbf{RQ5}: What is the influence of various hyperparameter settings on the predictive accuracy of \model?
\textbf{RQ6}: What visual explanations can be provided by the \model?

\vspace{-0.05in}
\subsection{Experimental Settings}
\subsubsection{\bf Datasets and Evaluation Protocols}
The experiments are conducted on both graph-based traffic prediction tasks and grid-based crime prediction tasks, utilizing five real-world datasets. The statistics of our experimental datasets are summarized in Table~\ref{tb:ds}. \\\vspace{-0.12in}

\begin{table}[t]
    \vspace{-0.1in}
    \centering
    \caption{Statistical information of the experimental datasets.}
    \vspace{-0.15in}
    \resizebox{.48\textwidth}{!}{\begin{tabular}{ccccccc} 
    \hline\hline
    Dataset   & Type  & Volume  & \# Interval & \# Nodes & \# Time Span      & \# Features  \\ 
  \hline
  PeMSD4    & Graph & Traffic & 5 min       & 307      & 01/2018 - 02/2018 & 1            \\
  PeMSD7    & Graph & Traffic & 5 min       & 883      & 05/2017 - 08/2017 & 1            \\
  PeMSD8    & Graph & Traffic & 5 min       & 170      & 07/2016 - 08/2016 & 1            \\
  NYC Crime & Grid  & Crime   & 1 day       & 256      & 01/2014 - 12/2015 & 4            \\
  CHI Crime & Grid  & Crime   & 1 day       & 168      & 01/2016 - 12/2017 & 4            \\
    \hline\hline
    \end{tabular}}
    \vspace{-0.2in}
    \label{tb:ds}
  \end{table}

\noindent\textbf{Traffic Prediction.} 
The model evaluation is firstly conducted using three widely used traffic datasets: PeMS04, PeMS07, and PeMS08~\cite{STGCN, ASTGCN, STSGCN, STGODE}. They were collected by the California Performance of Transportation (PeMS) and have a time interval of 5 minutes, covering different time ranges. To ensure a fair comparison, we split the datasets into training, validation, and testing sets in a 6:2:2 ratio. The evaluation of the models is performed using three metrics: \emph{Mean Absolute Error (MAE)}, \emph{Root Mean Squared Error (RMSE)}, and \emph{Mean Absolute Percentage Error (MAPE)}. \\\vspace{-0.12in}

\noindent\textbf{Crime Prediction.}
We also investigate the effectiveness of our model in spatio-temporal prediction using crime datasets: NYC Crime and CHI Crime. These datasets, collected from New York City and Chicago, respectively, capture crime incidents on a daily basis and are constructed with a spatial partition unit of $3km \times 3km$. Following the approach adopted in recent literature~\cite{ST-HSL, ST-SHN}, we generate the training and testing sets in a ratio of 7:1. In the training set, crime records from the last month are used for validation purposes. We utilize the MAE and MAPE as our evaluation metrics.

\noindent\textbf{Metrics for Explainability Analysis.}
Given the absence of ground-truths specifically designed for spatio-temporal explainability, we adopt metrics commonly used in the context of explainability for GNNs, namely \emph{Sparsity} and \emph{Fidelity}~\cite{XAI4GNN_survey}. To accommodate the spatio-temporal nature of our tasks, we make modifications to the \emph{Fidelity} metric, tailored for autoregressive tasks.
\begin{align}
  Fidelity+^{(s\backslash t)} = \frac{1}{Q} \sum_{i = 1}^{Q}(|f(\mathcal{G}_i^{(s\backslash t)}) - f(\mathcal{G}^{(s\backslash t)1-m_i}_i)|)
\end{align}
The modified \emph{Fidelity} metric, denoted as $Fidelity+^{(s\backslash t)}$, is utilized to measure the explainability of the Spatio-Temporal Graph (STG) framework. In this context, $Q$ represents the number of spatial and temporal graphs, $f$ represents the trained predictive spatio-temporal function, $\mathcal{G}_i^{(s\backslash t)}$ represents the i$^{th}$ original spatial$\backslash$temporal graph, and $m_i$ indicates the i$^{th}$ extracted explainable subgraph. Consequently, $\mathcal{G}^{(s\backslash t)1-m_i}$ refers to the masked spatial$\backslash$temporal graph based on the complementary subgraph structure $1-m_i$. Furthermore, the \emph{Sparsity} metric is redefined in the context of spatio-temporal graphs to capture the level of explainability.
\begin{align}
  Sparsity+^{(s\backslash t)} = \frac{1}{Q} \sum_{i = 1}^{Q}(1 - \frac{|m_i|}{|M_i|})
\end{align}
where $Sparsity+^{(s\backslash t)}$ indicates spatial/temporal \emph{Sparsity} of explainable subgraphs, $|m_i|$ and $|M_i|$ represent the number of important nodes based on the explainable subgraph and the original graph. 

\subsubsection{\bf Compared Baseline Methods}
We compare \model\ to methods, which can be categorized into two classes, to validate its performance in terms of both accuracy and explainability. \\\vspace{-0.12in}

\noindent\textbf{Predictive Accuracy:}
For the evaluation of our \model\ on traffic datasets, we compare it with 18 baselines that fall into 5 different categories. Similarly, for the evaluation on crime datasets, we adopt 12 baselines categorized into 5 different categories. \\\vspace{-0.12in}

\noindent \emph{Traffic Prediction:} 
(1) \textbf{Conventional Statistical Methods}: HA~\cite{HA}, VAR~\cite{HA};
(2) \textbf{Attention Methods}: ASTGCN~\cite{ASTGCN}, DSTAGNN~\cite{DSTAGNN};
(3) \textbf{Neural Differential Equation Models}: STG-ODE~\cite{STGODE}, STG-NCDE~\cite{STGNCDE};
(4) \textbf{GNN-based Methods}: DCRNN~\cite{DCRNN}, STGCN~\cite{STGCN}, GWN~\cite{GraphWaveNet}, STSGCN~\cite{STSGCN}, StemGNN~\cite{StemGNN}, AGCRN~\cite{AGCRN}, 
STFGNN~\cite{STFGNN}, Z-GCNETs~\cite{Z-GCNETs}, TAMP-S2GCNets~\cite{TAMP-S2GCNets}, 
GMSDR~\cite{GMSDR}, FOGS~\cite{FOGS}; 
(5) \textbf{Variant of \model}: \model-CGIB (\model\ with Conventional GIB)
\emph{Crime Prediction:}
(1) \textbf{Conventional Statistical Methods}: HA~\cite{HA}, SVM~\cite{SVM};
(3) \textbf{CNN-based Approach}: ST-ResNet~\cite{ST-ResNet};
(4) \textbf{Hybrid Spatio-Temporal Models}: ST-MetaNet~\cite{ST-MetaNet}, STDN~\cite{STDN}; (4) \textbf{Attention Methods}: DeepCrime~\cite{DeepCrime}, STtrans~\cite{STtrans}; (5) \textbf{GNN-based Models}: DCRNN~\cite{DCRNN}, STGCN~\cite{STGCN}, GMAN~\cite{GMAN}, ST-SHN~\cite{ST-SHN}, DMSTGCN~\cite{DMSTGCN}. \\\vspace{-0.12in}

\noindent\textbf{Predictive Explainability:}
To evaluate the predictive explainability of our approach, we employ four baselines, which can be grouped into two categories. (1) \textbf{Post-hoc Methods}: GNNExplainer~\cite{GNNExplainer}, PGExplainer~\cite{PGExplainer}, GraphMask~\cite{GraphMask}; (2) \textbf{Intrinsic Approach}: the \model-CGIB (\model\ with Conventional GIB).

\subsubsection{\bf Hyperparameter Settings}
Our \model\ is implemented using PyTorch and PyTorch Geometric library, with Adam optimizer, a learning rate of $1e^{-3}$, and a decay ratio of 0.5. We utilize two GAT layers with 16 heads for spatial and temporal encoding, with dimensions of 64 and 128, respectively. The prior probabilities $r^{(s)}$ and $r^{(t)}$ are scheduled with a decay ratio of 0.1 and a decay interval of 10 epochs. We employ an annealing strategy for $\lambda_1$ and $\lambda_2$ in the loss function to gradually change from 0 to 1 as the epoch increases. For traffic forecasting, we predict the next 12 time steps based on the past 12 time steps, while for crime prediction, we use 30 days of historical records to predict the next 1 day.

\begin{figure}[t]
  \centering
    \includegraphics[width=0.47\textwidth]{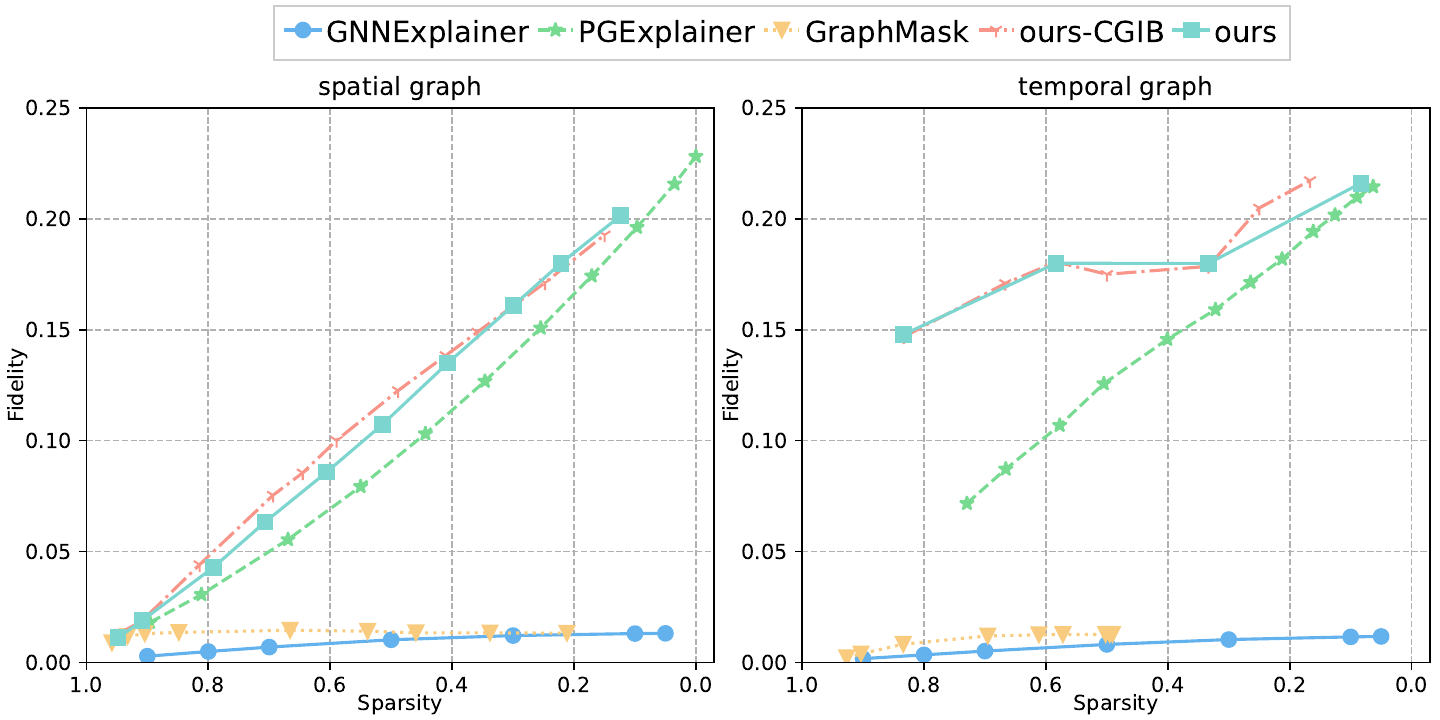}
    \vspace{-0.15in}
    \caption{Overall explainability comparison of \model.}
    \vspace{-0.25in}
  \label{fig:xai}
\end{figure}

\vspace{-0.1in}
\subsection{Prediction Accuracy Comparison (RQ1)}

\begin{figure}[t]
  \vspace{-0.05in}
  \centering
    
  \subfigure[node 144, from time step 505$\sim$ 793]{
      \centering
      \includegraphics[width=0.21\textwidth]{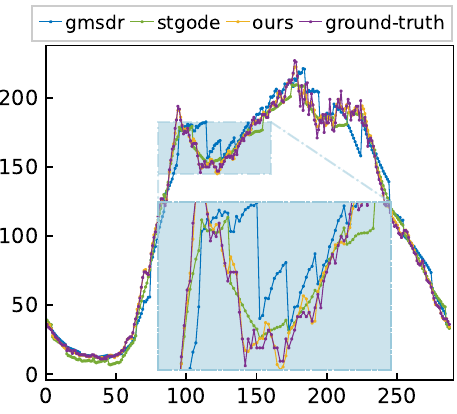}
  }\vspace{-3mm}
  \subfigure[node 193, from time step 0$\sim$ 288]{
    \centering
    \includegraphics[width=0.21\textwidth]{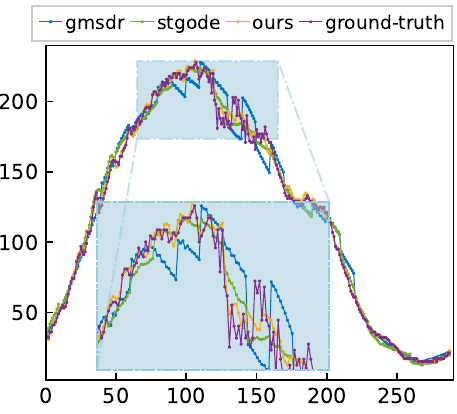}
  }
  \vspace{-1mm}
  \caption{Prediction visualization of our \model. }
  \vspace{-0.15in}

  \label{fig:pred_v1}
\end{figure}

Table~\ref{tb:cmp1} showcases the performance comparison results between our \model\ and state-of-the-art baselines on three traffic datasets. Additionally, Table~\ref{tb:cmp2} presents the comparison results for crime prediction, emphasizing the best-performing model in each dataset. Based on these results, we have the following observations:
\begin{itemize}[leftmargin=*]
  \item \textbf{Overall Superiority of \model.} Our \model\ consistently outperforms state-of-the-art baselines in both tasks. This is attributed to its effective architecture, utilizing an STG attentive encoder and decoder with a position-aware fusion layer. Additionally, the incorporation of explainable structure-distilled Graph Information Bottleneck (GIB) helps filter out irrelevant information and noise, improving accuracy and interpretability. \\\vspace{-0.12in}
  

  \item \textbf{Comparing to State-of-the-arts.} Compared to attention-based models like ASTGCN, DSTAGNN, DeepCrime, and STtrans, our \model\ achieves significant improvements in predictive performance. The explainable GIB principle filters task-irrelevant structural correlations, allowing attentive information to propagate over influential subgraphs. The spatio-temporal GIB demonstrates generalization and robustness, extracting task-relevant information from sparse crime data. The performance gap with GNN-based approaches (FOGS, GMSDR, TAMP-S2GCNets, Z-GCNETs, DMSTGCN, ST-SHN, GMAN) highlights the effectiveness of using graph attention mechanisms to alleviate the over-smoothing effects while modeling complex spatial and temporal correlations. Furthermore, comparing our \model\ with the variant \model-CGIB, which surpasses most baselines, further confirms the effectiveness of our framework. The GIB principle, instantiated by unified STG attention networks, plays a crucial role in improving performance. \\\vspace{-0.12in}
  

  \item \textbf{Visualization of predictions.} We further visualize the predictive results on PEMS04, demonstrating the comparison between our \model\ and two competitive baselines, namely STG-ODE and GMSDR, along with the ground-truth results. The visual comparison, depicted in Figure~\ref{fig:pred_v1}, highlights the superiority of our \model. It excels in predicting inflection points that involve sharp jitter changes due to its capability to filter out task-irrelevant information, capturing essential spatio-temporal dynamics, and providing more accurate results.

\end{itemize}
\begin{table}[t]
    \vspace{-0.15in}
    \centering
    \caption{Performance comparison on NYC, CHI crimes.}
    \vspace{-0.15in}
    \resizebox{0.37\textwidth}{!}{\begin{tabular}{c|cc|cc} 
    \hline
    \multirow{2}{*}{Model} & \multicolumn{2}{c|}{NYC Crime}    & \multicolumn{2}{c}{CHI Crime}      \\ 
    \cline{2-5}
                           & MAE             & MAPE            & MAE             & MAPE             \\ 
    \hline
    HA                  & 1.0765          & 0.6196          & 1.2616          & 0.5894           \\
    SVM                    & 1.2805          & 0.6863          & 1.3622          & 0.5992           \\
    ST-ResNet              & 0.9755          & 0.5453          & 1.1014          & 0.5294           \\
    DCRNN                  & 0.9638          & 0.5569          & 1.0885          & 0.5260           \\
    STGCN                  & 0.9538          & 0.5451          & 1.0970          & 0.5283           \\
    STtrans                & 0.9640          & 0.5584          & 1.0817          & 0.5179           \\
    DeepCrime              & 0.9429          & 0.5496          & 1.0801          & 0.5166           \\
    STDN                   & 0.9993          & 0.5762          & 1.1245          & 0.5480           \\
    ST-MetaNet             & 0.9572          & 0.5620          & 1.0913          & 0.5225           \\
    GMAN                   & 0.9587          & 0.5575          & 1.0752          & 0.5166           \\
    ST-SHN                 & 0.9280          & 0.5373          & 1.0689          & 0.5116           \\
    DMSTGCN                & 0.9293          & 0.5485          & 1.0736          & 0.5175           \\
    \model-CGIB                & 0.9287          & 0.5394          & 1.0701          & 0.5143           \\
    \textbf{\model}        & \textbf{0.9095} & \textbf{0.5154} & \textbf{1.0307} & \textbf{0.5016}  \\
    \hline
    \end{tabular}}
    \label{tb:cmp2}
    \vspace{-0.15in}
    \end{table}

\vspace{-0.1in}
\subsection{Model Explainability Evaluation (RQ2)}
In this subsection, we quantitatively analyze the spatio-temporal explainability of our \model. We use modified metrics, namely "Sparsity" and "Fidelity," to evaluate the spatial and temporal graphs. The comparison results on PEMS04 are shown in Figure~\ref{fig:xai}. To ensure a fair comparison, we employ \post\ frameworks to explain STGNN models using the same STG encoder and decoder. Higher scores in "Sparsity" and "Fidelity" indicate better predictive explainability, aiming to extract smaller, impactful spatio-temporal subgraphs. The \model\ framework, incorporating explainable information bottleneck, achieves the best explainable performance compared to state-of-the-art approaches. This validates the effectiveness of injecting explainability into our unified STGNN architecture with the IB principle. Among the \post\ methods, PGExplainer outperforms others for STGNN models, providing effective global explanations. However, the performance gap between our \intr\ models with \model\ framework and \post\ methods highlights our inherent superiority in providing faithful explanations for spatio-temporal GNN architecture.
 
\vspace{-0.05in}
\subsection{Ablation Study (RQ3)}
\begin{figure}[t]
  \vspace{-0.1in}
  \centering
  \subfigure[On PEMS4]{
      \centering
      \includegraphics[width=0.4\textwidth]{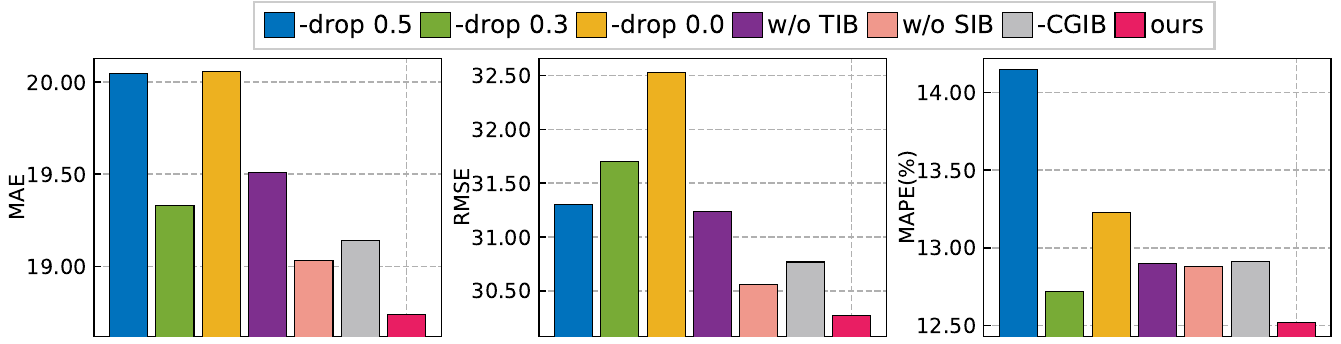}
  }\vspace{-3mm}
  \vspace{-3mm}
  \subfigure[On PEMS8]{
    \centering
    \includegraphics[width=0.4\textwidth]{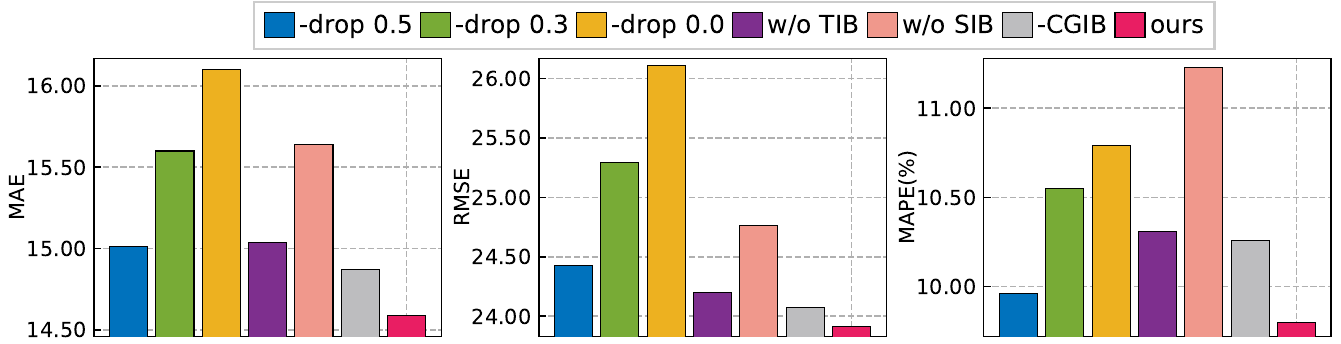}
  }
  \vspace{-1mm}
  \caption{Ablation experiments of our \model.}

  \label{fig:ablation}
  \vspace{-0.2in}
\end{figure}

We investigate the effectiveness of the proposed modules by designing variants of our \model: i) "-CGIB": We replace the explainable GIB principle with the conventional one to compare different ways of controlling bottleneck information. ii) "-w/o SIB" and "-w/o TIB": We remove the explainable GIB-based GAT encoder in spatial and temporal modeling, respectively, and use canonical GAT instead. iii) "-drop 0.5", "-drop 0.3", and "-drop 0.0": We randomly drop edges of the spatio-temporal graphs with different probabilities, instead of utilizing the explainable spatio-temporal GIB. We analyze the results on PEMS04 and 08, which are shown in Figure~\ref{fig:ablation}. Through these experiments, we make the following discoveries:
\begin{itemize}[leftmargin=*]

  \item Regarding the "-CGIB" variant, the performance improvements achieved by our \model\ demonstrate the superiority of the explainable structure-distilled GIB over the conventional one in controlling the flow of structural information during inference. This is attributed to the significance and sensitivity of graph structure in graph neural network models.
  

  \item The application of explainable GIB demonstrates its effectiveness in capturing important spatial and temporal dependencies, as evidenced by the variants "-w/o SIB" and "-w/o TIB". The influence of spatial and temporal GIB on the framework's performance depends on the credibility and noise levels present in the original spatial and temporal graph structures. In this regard, we argue that the fully connected temporal graph often contains more task-irrelevant structural noise that needs to be filtered out.


  \item When examining the variants "-drop 0.5", "-drop 0.3", and "-drop 0.0", a noticeable performance gap becomes apparent. This gap arises because random edge dropping cannot differentiate between task-relevant and task-irrelevant edges in the graph structure. It is worth noting that "-drop 0.5" and "-drop 0.3" outperform "-drop 0.0" in terms of better generalization. This outcome validates the necessity of edge dropping and motivates us to develop more accurate and efficient strategies for edge drop in the future.
  
\end{itemize} 

\begin{figure}[t]
  \vspace{-0.15in}
  \centering
    
  \subfigure[On NYC Crime]{
      \centering
      \includegraphics[width=0.45\textwidth]{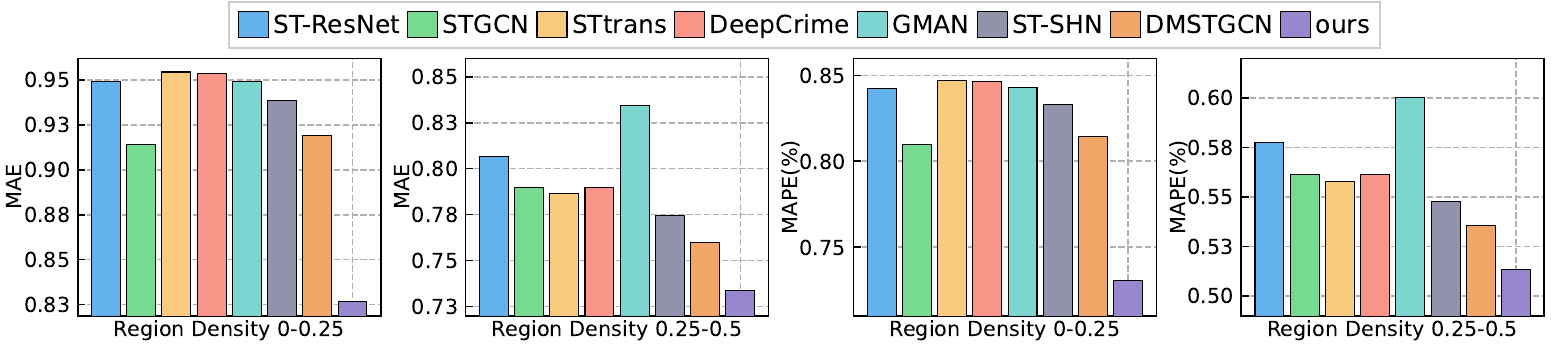}
  }\vspace{-3mm}
  \subfigure[On CHI Crime]{
    \centering
    \includegraphics[width=0.45\textwidth]{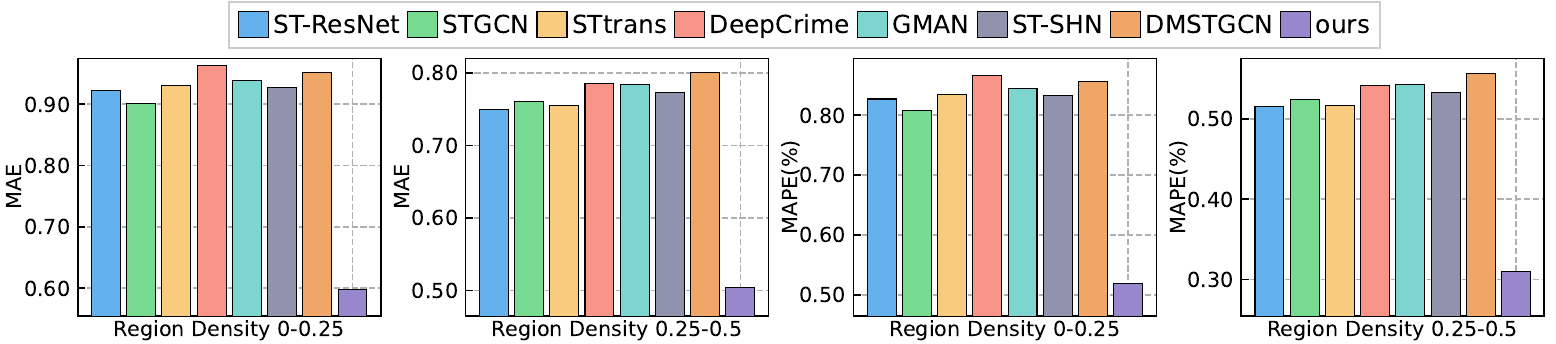}
  }
  \vspace{-0.15in}
  \caption{Performance comparison on sparse regions.}

  \label{fig:sparsity}
  \vspace{-0.1in}
\end{figure}

\vspace{-0.1in}
\subsection{Generalization and Robustness Study (RQ4)}
The inherent ability of GIB to extract task-relevant and prediction-influential spatio-temporal information allows us to further validate the generalization and robustness of our \model. To achieve this, we address two specific data quality issues.\\\vspace{-0.12in}

\noindent\textbf{Performance \wrt\ Data Missing.} 
In real-world spatio-temporal scenarios, data missing challenges often arise due to sensor failure and privacy policies. To assess the performance of our \model\ in such cases, we randomly drop traffic volumes on each node with proportions of 10\%, 30\%, and 50\% for traffic prediction. It is important to note that the data drop between nodes is independent. The results on PEMS04 are presented in Table~\ref{tb:missing}, where "-" indicates that the model fails in this situation. The results demonstrate that our \model\ is capable of filtering out random noise introduced by data drop, showcasing its generalization and robustness. We compare our \model\ with three competitive baselines, namely STGODE, GMSDR, and STG-NCDE. We observe that the performance of STGODE and GMSDR sharply decreases with an increasing proportion of dropped data. However, our \model\ shows its robustness by achieving better performance in adapting to data missing scenarios compared to STG-NCDE. \\\vspace{-0.12in}

\begin{table}[t]
  \centering
  \caption{Performance comparison against data missing.}
  \vspace{-0.1in}
  \resizebox{0.49\textwidth}{!}{\begin{tabular}{c|ccc|ccc|ccc} 
  \hline
  \multirow{3}{*}{model} & \multicolumn{9}{c}{PEMS04}                                                                                \\ 
  \cline{2-10}
                         & \multicolumn{3}{c|}{missing 10\%} & \multicolumn{3}{c|}{missing 30\%} & \multicolumn{3}{c}{missing 50\%}  \\ 
  \cline{2-10}
                         & MAE   & RMSE  & MAPE              & MAE   & RMSE  & MAPE              & MAE   & RMSE  & MAPE              \\ 
  \hline
  STGODE                 & 23.97 & 35.41 & 19.13             & 45.02 & 59.48 & 29.54             & ~-~   & ~-~   & ~-~               \\
  GMSDR                  & 21.69 & 34.06 & 13.81             & 25.02 & 38.45 & 15.01             & 103.01& 131.64& 47.31                  \\
  ST-NCDE                & 19.36 & 31.28 & 12.79             & 19.40 & 31.30 & 13.04             & 19.98 & 32.09 & 13.48             \\
  \model\                 & 19.12 & 30.84 & 12.61             & 19.34 & 31.20 & 12.86             & 19.92 & 32.05 & 13.27             \\
  \hline
  \end{tabular}}
  \label{tb:missing}
  \vspace{-0.1in}
  \end{table}

\noindent\textbf{Performance \wrt\ Data Sparsity.} 
In practical scenarios, e.g., crime prediction, spatio-temporal signals across the observed space often exhibit sparsity, with many regions or nodes having zero values. This poses a challenge for achieving better generalization and robustness of the model. To address this, we categorize regions in crime prediction tasks based on historical region density. We compare the predictive results of our \model\ with baselines on density ranges "0-0.25" and "0.25-0.5", as depicted in Figure~\ref{fig:sparsity}. The notable performance gap underscores the capability of our \model\ framework to extract task-relevant information from sparse data, resulting in improved predictive performance. This is particularly noteworthy as our \model\ outperforms methods specifically tailored for crime prediction, e.g., STtrans and ST-SHN.

\subsection{Hyperparameter Investigation (RQ5)}
\begin{figure}[t]
  \centering
    \includegraphics[width=0.42\textwidth]{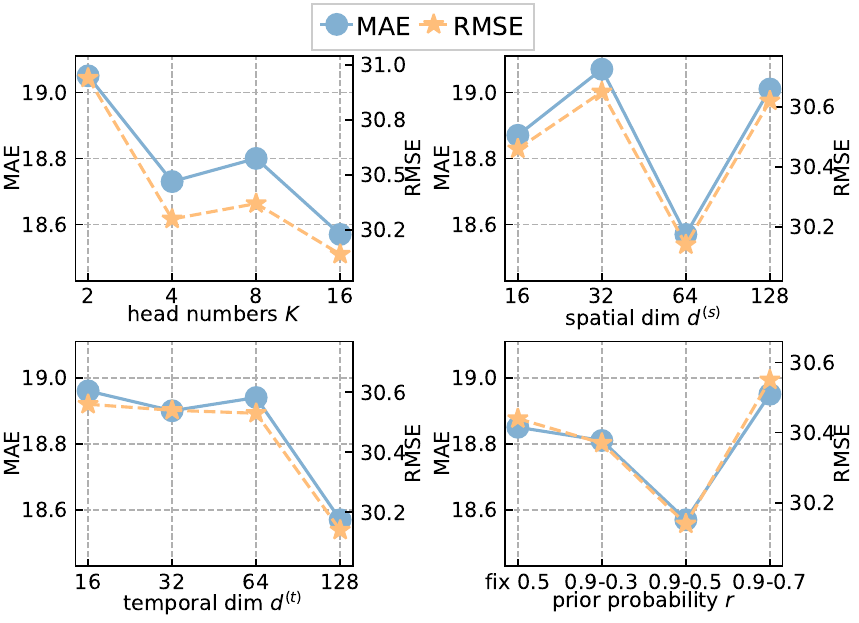}
    \vspace{-0.15in}
    \caption{Hyperparameter study of the proposed \model.}
    \vspace{-0.15in}
  \label{fig:para}
\end{figure}

We are conducting a hyperparameter investigation by varying specific hyperparameters while keeping others at their default values. We focus on four significant hyperparameters: head numbers ($K$), prior probability ($r$), spatial dimension ($d^{(s)}$), and temporal dimension ($d^{(t)}$). The results of our experiments on PEMS04 are displayed in Figure~\ref{fig:para}. Here are our detailed experiments and observations: i) \textbf{Head numbers} ($K$): We vary the number of heads in the spatio-temporal GAT encoder from the range of ${2, 2^2, 2^3, 2^4}$. We find that the model with $2^4$ heads achieves the best performance. Increasing the number of heads enables the model to capture spatio-temporal correlations from multiple dimensions. ii) \textbf{Spatial and temporal dimensions} ($d^{(s)}$, $d^{(t)}$): We search for the optimal spatial and temporal dimensions in the spatio-temporal GAT encoder within the range of ${2^4, 2^5, 2^6, 2^7}$. We find that $d^{(s)} = 2^6$ and $d^{(t)} = 2^7$ serve as the best settings. iii) \textbf{Prior probability} ($r$): The prior probability $r$ represents the spatio-temporal prior probability in Equation~\ref{tb:ib_loss}. We explore the search range of {"fix 0.5", "0.9-0.3", "0.9-0.5", "0.9-0.7"}, where "fix 0.5" indicates fixing $r$ at 0.5, and the last three options involve controlling $r$ to decay from 0.9 as the epoch increases.

\begin{figure}[t]
  \centering
    
  \subfigure[spatio-temporal pattern explanations]{
      \centering
      \includegraphics[width=0.22\textwidth]{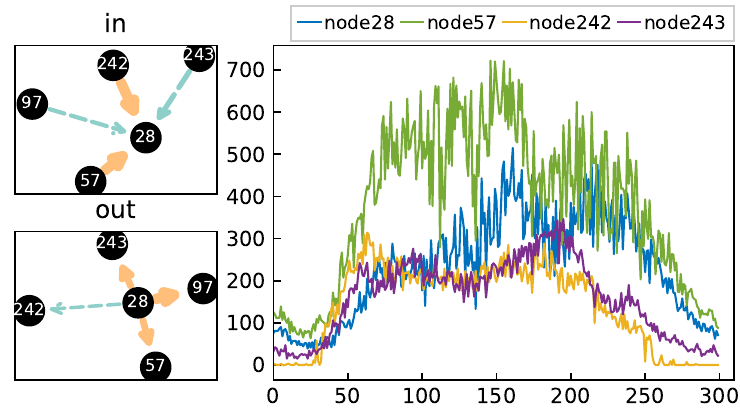}
      \includegraphics[width=0.22\textwidth]{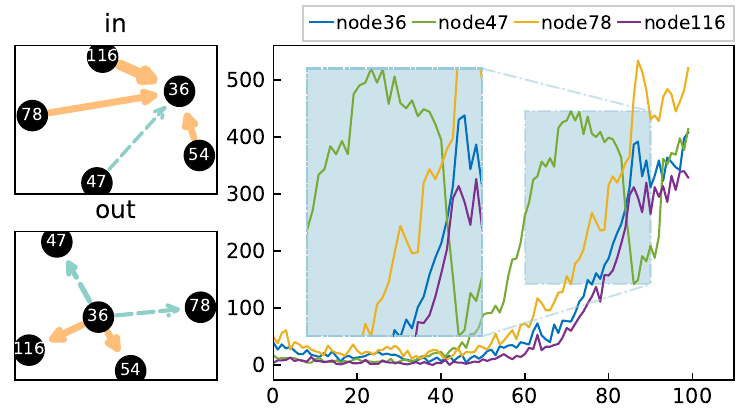}
  }\vspace{-2mm}
  \subfigure[spatial semantics explanations]{
    \centering
    \includegraphics[width=0.44\textwidth]{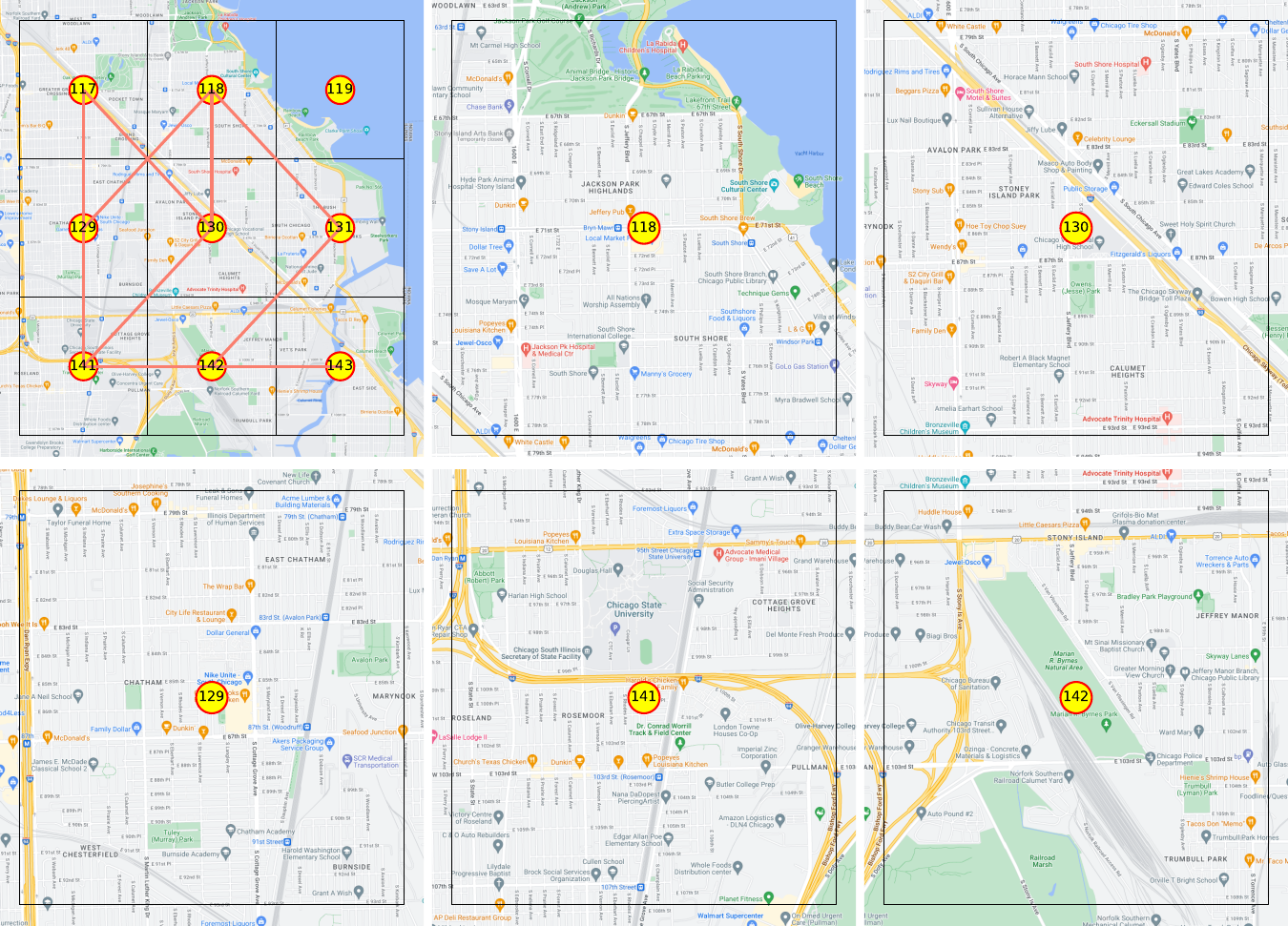}
  }
  \vspace{-0.1in}
  \caption{Case study of our \model. }
  \label{fig:case}
  \vspace{-0.2in}
\end{figure}

\subsection{Model Interpretation Case Study (RQ6)}
We conduct an investigation into the explanations provided by our \model\ in identifying important subgraphs and analyzing their spatio-temporal patterns. To achieve this, we employ a method of obtaining explainable subgraphs by discarding edges with low edge weights. Subsequently, we employ extensive visualization techniques to gain a deeper understanding of the relationships. \\\vspace{-0.12in}

\noindent\textbf{Spatio-Temporal Pattern Explanations:} 
In the left side of Figure~\ref{fig:case} (a), our \model\ identifies node 28 as being more related to node 57 compared to nodes 242 and 243. This relationship is also apparent in the adjacent time series diagram, where node 28 and node 57 exhibit similar time trend patterns with comparable peaks and valleys. The thickness of the arrow connecting the nodes represents the edge weights encoded by the GIB principle. Furthermore, in the right side of Figure~\ref{fig:case} (a), we observe that node 47 is considered to be weakly correlated with node 36 and other nodes. This is evident from the highlighted peaks, valleys, and rising time points in the adjacent time series plot. Both figures demonstrate that our \model\ provides explanations that accurately reflect spatio-temporal trend patterns across time and locations. \\\vspace{-0.12in}

\noindent\textbf{Spatial Semantics Explanation:} 
Due to the unavailability of coordinate information in the PEMS04, PEMS07, and PEMS08 datasets, we focus on exploring the precise semantic information provided by explanations on the CHI Crimes dataset. Figure~\ref{fig:case} (b) visually present our findings, where interconnected regions exhibit similar regional functionality, particularly in terms of shared Point of Interest (POI) information. For example, regions 118 and 129, as well as regions 141 and 142, display comparable POI characteristics, implying functional resemblance. Conversely, region 119 stands out as relatively isolated due to its primarily oceanic nature and lack of substantial POI information. These findings underscore the significance of explainability and the prediction effectiveness.

\section{Conclusion}
\label{sec:conclusion}
In this study, we emphasize the significance of explainability in spatio-temporal graph neural networks. To address this, we propose a novel framework called \model\ that not only predicts future spatio-temporal signals accurately but also provides transparent explanations. Our framework incorporates GIB-based structure distillation with an explainable objective and employs variational approximation for tractability. Additionally, we introduce a unified STG encoder and decoder that generate explainable, generalizable, and robust STG representations. Through extensive experiments, we demonstrate the superiority of our \model\ in terms of predictive accuracy, explainability, generalization, and robustness. Our results surpass existing state-of-the-art methods in both predictive accuracy and explainability. In future work, we plan to investigate effective approaches for integrating explainability into global spatial information propagation mechanisms, such as hypergraph neural networks, using an \intr\ explainable approach.



\clearpage
\balance
\bibliographystyle{ACM-Reference-Format}
\bibliography{sample-base}



\end{document}